\documentclass[11pt]{article}

\usepackage{acl}

\usepackage{times}
\usepackage{latexsym}

\usepackage[T1]{fontenc}

\usepackage[utf8]{inputenc}

\usepackage{microtype}

\usepackage{inconsolata}

\usepackage{graphicx}

\usepackage{algorithm}

\usepackage{booktabs}

\usepackage{caption}

\usepackage{multirow}

\usepackage{framed}
\usepackage[most]{tcolorbox}

\usepackage{url}

\usepackage{amsmath}

\usepackage{amssymb}
\usepackage[noend]{algpseudocode}

\usepackage{enumitem}

\usepackage{xspace}
\newcommand{\methodname}{RRPO\xspace}
\newcommand{\methodfullname}{ReRanking Preference Optimization\xspace}

\title{Optimizing RAG Rerankers with LLM Feedback \\
via Reinforcement Learning}

\author{
  Yuhang Wu\textsuperscript{1}\thanks{Equal contribution.} \quad
  Xiangqing Shen\textsuperscript{2}\footnotemark[1] \quad
  Fanfan Wang\textsuperscript{1} \quad
  Cangqi Zhou\textsuperscript{1} \\
  \textbf{Zhen Wu}\textsuperscript{3} \quad
  \textbf{Xinyu Dai}\textsuperscript{3} \quad
  \textbf{Rui Xia}\textsuperscript{2}\thanks{Corresponding author.} \\
  \textsuperscript{1}School of Computer Science and Engineering, Nanjing University of Science and Technology, China \\
  \textsuperscript{2}School of Intelligence Science and Technology, Nanjing University, China \\
  \textsuperscript{3}School of Artificial Intelligence, Nanjing University, China \\
  \texttt{\{yhangwu, ffwang, cqzhou\}@njust.edu.cn} \\
  \texttt{\{xqshen, wuz, daixinyu, rxia\}@nju.edu.cn}
}

\begin{document}

\maketitle

\begin{abstract}
Rerankers play a pivotal role in refining retrieval results for Retrieval-Augmented Generation.
However, current reranking models are typically optimized on static human annotated relevance labels in isolation, decoupled from the downstream generation process.
This isolation leads to a fundamental misalignment:
documents identified as topically relevant by information retrieval metrics often fail to provide the actual utility required by the LLM for precise answer generation.
To bridge this gap, we introduce \methodfullname (\methodname)\footnote{Our code is publicly available at \url{https://github.com/NUSTM/RRPO}}, a reinforcement learning framework that directly aligns reranking with the LLM's generation quality.
By formulating reranking as a sequential decision-making process, \methodname optimizes for context utility using LLM feedback, thereby eliminating the need for expensive human annotations.
To ensure training stability, we further introduce a reference-anchored deterministic baseline.
Extensive experiments on knowledge-intensive benchmarks demonstrate that \methodname significantly outperforms strong baselines, including the powerful list-wise reranker RankZephyr.
Further analysis highlights the versatility of our framework: it generalizes seamlessly to diverse readers (e.g., GPT-4o), integrates orthogonally with query expansion modules like Query2Doc, and remains robust even when trained with noisy supervisors.
\end{abstract}

\section{Introduction}
\label{sec:introduction}

Retrieval-Augmented Generation (RAG)~\cite{10.5555/3495724.3496517,borgeaud2022improving,izacard-grave-2021-leveraging,gao2023retrieval} has emerged as a powerful paradigm to enhance LLMs by dynamically incorporating external information, thereby improving factual accuracy and contextual relevance for knowledge-intensive tasks.
In a typical pipeline, the system first retrieves documents based on a user query and then passes them to an LLM Reader for response generation.
Consequently, the overall performance depends heavily on the quality of these retrieved documents.
To optimize this, a reranking model (reranker) is often introduced to refine the initial document set.
By prioritizing the most relevant documents, the reranker ensures the LLM receives high-quality context, leading to more precise and coherent answers.

\begin{figure*}[t]
  \centering
  \includegraphics[width=1\textwidth]{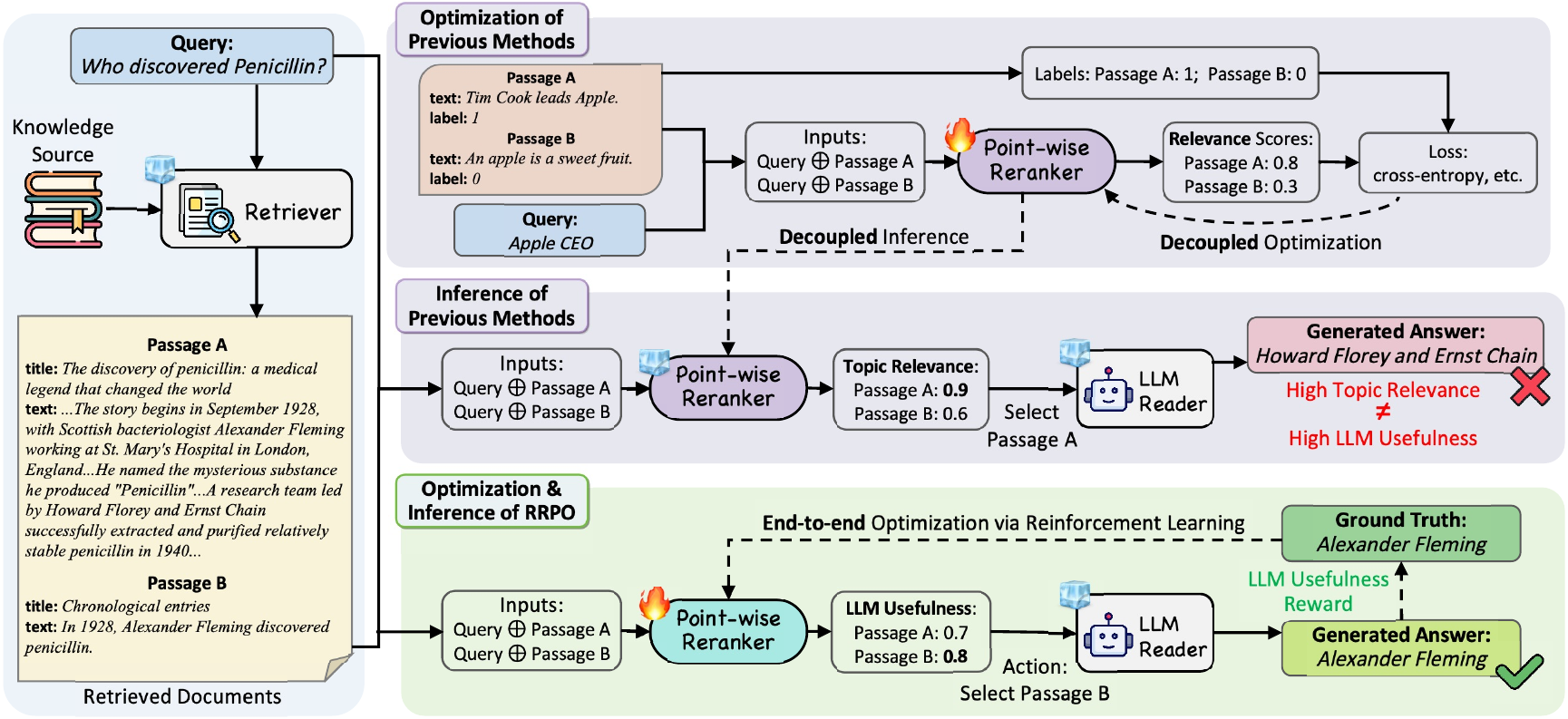}
  \caption{Comparison of previous methods with our proposed \methodname framework.}
  \label{fig:intro_overview}
\end{figure*}

However, a fundamental disconnect persists in current RAG architectures: reranker optimization is typically isolated from the RAG pipeline.
Most rerankers are optimized under a supervised paradigm using static Information Retrieval metrics (e.g., NDCG), a process that is entirely decoupled from the downstream LLM.
As shown in Figure~\ref{fig:intro_overview}, this creates a critical misalignment:
the notion of relevance learned by the reranker does not necessarily translate to the usefulness required by the LLM to generate precise answers~\cite{lee-etal-2025-shifting}.
For instance, given the query ``Who discovered penicillin?'', a standard reranker might favor a long historical article (high topic relevance), whereas the LLM might benefit more from a concise timeline entry (high LLM usefulness).
By treating the reranker as an independent module rather than an integrated component of the RAG pipeline, standard approaches
may not explicitly 
optimize for the system's ultimate objective---generation quality.

To bridge this gap, we introduce \methodfullname (\methodname), a reinforcement learning framework that directly optimizes the reranker within the RAG pipeline using the LLM's generation quality as the reward signal.
Unlike supervised approaches that rely on static, human-annotated relevance labels, \methodname leverages the downstream LLM itself to provide scalable supervision, thereby eliminating the need for expensive manual annotation.
\methodname reframes reranking as a sequential decision-making process, allowing for end-to-end alignment between retrieval and generation.
To address the instability caused by discrete, high-variance LLM rewards, we introduce a reference-anchored deterministic baseline.
This component effectively stabilizes training by compelling the reranker to prioritize document ``usefulness'' over mere topic ``relevance'', avoiding the computational complexity and instability associated with training a separate critic network.

We validate the effectiveness of \methodname through extensive experiments on standard knowledge-intensive benchmarks.
Empirical results demonstrate that \methodname achieves consistent improvements by explicitly optimizing the reranker to align with reader needs, significantly outperforming strong supervised baselines.
Notably, our method surpasses state-of-the-art list-wise LLM rerankers, RankZephyr~\cite{pradeep2023rankzephyr}.
This confirms that aligning the reranker with the reader's specific needs yields good context utility.

Beyond performance gains, our analysis highlights three critical properties that distinguish \methodname as a robust solution.
First, it exhibits strong generalization: the learned policy not only handles complex multi-hop reasoning but also transfers seamlessly to diverse LLM Readers, effectively enhancing even proprietary models like GPT-4o.
Second, it offers seamless integration: functioning as a ``plug-and-play'' module, it complements advanced retrieval strategies and delivers additive gains when combined with query expansion methods like Query2Doc~\cite{wang-etal-2023-query2doc}.
Finally, it demonstrates training robustness: \methodname maintains its effectiveness even when optimized using feedback from smaller, noisy supervisors, significantly reducing training costs.

\section{Related Work}
\label{sec:related-work}

\subsection{Retrieval-Augmented Generation}

Retrieval-Augmented Generation (RAG) enhances LLMs by incorporating external information to mitigate hallucinations and knowledge obsolescence~\cite{ji2023survey,guu2020retrieval,10.5555/3495724.3496517}. While early frameworks focused on dynamic knowledge provisioning for QA and dialogue~\cite{gao2023retrieval,karpukhin-etal-2020-dense,izacard2023atlas,shuster-etal-2021-retrieval-augmentation}, subsequent research has refined the generation process. Key advancements include instruction-tuning generators to better utilize context~\cite{ma2024fine,zhu-etal-2024-inters,muennighoff2024generative,liu2024chatqa,lin2023ra} and incorporating self-reflection mechanisms like Self-RAG~\cite{asai2024self}. Furthermore, sophisticated retrieval strategies such as iterative, adaptive, or active retrieval have been developed to handle complex queries~\cite{trivedi-etal-2023-interleaving,jiang2023active,jeong-etal-2024-adaptive,xu2024recomp}.

\subsection{Rerankers in RAG}

Rerankers serve as a critical refinement stage between retrieval and generation. Early methods employed BERT-based encoders for pointwise or pairwise relevance estimation~\cite{nogueira2019passage,nogueira2019multi}. Recent approaches explicitly leverage LLM capabilities, ranging from zero-shot listwise prompting~\cite{sun-etal-2023-chatgpt} or discrete prompt optimization~\cite{cho2023discrete}, to fine-tuning open-source LLMs on ranking data, as seen in RankZephyr~\cite{pradeep2023rankzephyr,ma2024fine}. Other works have explored unsupervised risk minimization~\cite{yuan2024improving} or utilized attention mechanisms for efficiency~\cite{gangi-reddy-etal-2024-first, chen2024attention}. Most recently, research has expanded into context-aware and agentic paradigms.
EBCAR~\cite{yuan2025embedding} utilizes embeddings to capture cross-passage interactions, while REARANK~\cite{zhang2025rearank} formulates reranking as a reasoning agent.

\subsection{Reinforcement Learning for RAG}
\label{subsec:rl-for-rag}

Reinforcement Learning (RL) has been widely adopted to align RAG components with end-task performance. Prior work has applied RL to optimize query rewriting~\cite{wang2023promptagent}, incorporate reasoning traces via ReAct~\cite{yao2023react}, or facilitate multi-step browsing~\cite{nakano2021webgpt}. A growing trend involves aligning retrieval with generation via preference optimization. Methods like DynamicRAG~\cite{sun2025dynamicragleveragingoutputslarge}, KnowPO~\cite{Zhang_Xu_Xiao_Zhu_Jiang_Chu_Zhao_Wang_2025}, and DPA-RAG~\cite{dong2025understand} apply preference optimization (e.g., DPO) to fine-tune models, while others explore multi-agent reinforcement learning architectures~\cite{chen2025improving}. While recent progress has been driven by DPO-based alignments and multi-agent frameworks, \methodname explores an alternative paradigm by employing a sequential decision-making formulation to explicitly capture the combinatorial utility of document sets.

\section{Methodology}
\label{sec:methodology}

We propose \methodname, a framework that fundamentally redefines document reranking as a sequential decision-making process.
This approach is designed to bridge the misalignment between the relevance scores predicted by traditional rerankers and the actual usefulness of documents for the downstream LLM.
Unlike supervised methods that rely on static labels, \methodname leverages Reinforcement Learning (RL) to optimize document selection based on direct feedback from the frozen LLM Reader.
This allows for end-to-end alignment of the retrieval pipeline with the final generation quality.
In this section, we formulate the reranking problem as a Markov Decision Process (MDP) and detail the components of our training algorithm, highlighting our reference-anchored baseline.

\subsection{Task Formulation}
\label{subsec:task-formulation}

Given a user query $q$ and an initial candidate set of $N$ documents $\mathcal{D}=\{d_1, \dots, d_N\}$ retrieved from a corpus, the goal of the reranker is to select an ordered subset of $k$ documents $(k < N)$ that maximizes the quality of the response generated by a fixed LLM Reader.

We formalize this selection process as a finite-horizon Markov Decision Process (MDP) defined by the tuple $(\mathcal{S}, \mathcal{A}, P, R)$:

\textbf{State $\left(s_t\right)$}:
At time step $t$, where $t \in \{1, \dots, k\}$, the state $s_t$ is defined as the set of candidate documents that have not yet been selected.
The initial state $s_1=\left\{d_1, d_2, \ldots, d_N\right\}$ is the set of top-$N$ documents returned by the initial retrieval module.

\textbf{Action $\left(a_t\right)$}:
In state $s_t$, an action $a_t$ denotes selecting one document $d_{c_t}$ from the current set of documents $s_t$. Here, $c_t$ is the index of this document in the original list of $N$ documents. The action space $\mathcal{A}_t$ varies at each time step:
\begin{equation}
\mathcal{A}_t=\left\{\text {select} \, d_i \mid d_i \in s_t\right\} .
\end{equation}

\textbf{Transition Dynamics $\left(P(s_{t+1} \mid s_t, a_t)\right)$}:
The state transitions are deterministic. Upon executing action $a_t$ (i.e., selecting document $d_{c_t}$) in state $s_t$, the system transitions to the next state $s_{t+1}$, which is obtained by removing the selected document $d_{c_t}$ from $s_t$ :
\begin{equation}
s_{t+1} = s_t \setminus \{d_{c_t}\} .
\end{equation}

\textbf{Reward $\left(r_t\right)$}:
After executing action $a_t$ at each time step $t$, the agent receives a reward $r_t$. This reward reflects the contribution of the set of documents selected up to the current time step $t$, $\left\{d_{c_1}, \ldots, d_{c_t}\right\}$, to the generation of a high-quality response by the downstream LLM Reader. The downstream LLM Reader generates a specific reward value based on an evaluation metric $R_{lm}$. 

\subsection{ReRanking Preference Optimization}
\label{subsec:rrpo}

We detail our \methodfullname framework in three aspects: Agent and Policy Network, Optimization objective, and Reference-anchored Deterministic Baseline.

\subsubsection{Agent and Policy Network}
In this framework, the agent is implemented by a parameterized reranking model $f_\theta$ with parameters $\theta$.
The agent's behavior is defined by a policy $\pi_\theta\left(a_t \mid s_t\right)$.
We realize our policy $\pi_\theta$ through a pointwise reranking model $f_\theta$.
For a given query $q$ and each document $d_i(i=1, \ldots, N)$ in the initial candidate list, the reranker model $f_\theta$ computes a relevance score:
\begin{equation}
\label{eq:f_theta}
\text{score}_i=f_\theta\left(q, d_i\right) ,
\end{equation}
where the scores for all $N$ candidate documents are then converted into an initial probability distribution $\left\{p_1, \ldots, p_N\right\}$ using the SoftMax function, where $p_i$ represents the initial probability that document $d_i$ is considered relevant:
\begin{equation}
p_i=\frac{\exp \left(\text{score}_i\right)}{\sum_{j=1}^N \exp \left(\text{score}_j\right)} .
\end{equation}
At each Reinforcement Learning time step $t$, when the agent is in state $s_t$ (the set of remaining documents), the probability of selecting action $a_t$ (i.e., choosing document $d_{c_t} \in s_t$), $\pi_\theta\left(a_t \mid s_t\right)$ is defined as the proportion of this document's initial probability $p_{c_t}$ to the sum of initial probabilities of the currently remaining documents:
\begin{equation}
\begin{split}
\pi_\theta&(a_t=\text{select} \, d_{c_t} \mid s_t) \\
&= \frac{p_{c_t}}{\sum_{d_j \in s_t} p_j}
= \frac{p_{c_t}}{1-\sum_{l=1}^{t-1} p_{c_l}} .
\end{split}
\end{equation}

\textbf{Reward Calculation}:
The sequence of selected documents $\left\{d_{c_1}, \ldots, d_{c_t}\right\}$ is combined with the original query $q$ and an instruction INST, and then input to the LLM Reader to generate a response:
\begin{equation}
\text{response}_t=\text{Reader}_{\text {LLM}}\left(\text {INST}, q, d_{c_1}, \ldots, d_{c_t}\right) .
\end{equation}
An evaluation metric $R_{lm}$ is used to compute a score between the generated response $\text {response}_t$ and the ground truth answer ans, which serves as the reward for that time step:
\begin{equation}
\label{eq:compute_reward}
r_t=R_{lm}\left(\text{ans}, \text{response}_t\right) .
\end{equation}
The specific settings for the instruction INST and the evaluation metric $R_{lm}$ are dataset-dependent and will be detailed in the Experiments section.

\subsubsection{Objective Function}
\label{subsubsec:objective-function}

In practice, we found it necessary to constrain the magnitude of updates to the reranker model to prevent excessive oscillations in the training process that can result from large policy gradient updates.
Therefore, drawing inspiration from the Proximal Policy Optimization (PPO) algorithm, we define our specific training objective as
\begin{equation}
\label{eq:ppo-objective}
\begin{split}
    J(\theta) = \mathbb{E}_{\tau \sim \pi_\theta} \Bigg[ & \frac{1}{k} \sum_{t=1}^{k} \min \bigg( \rho_t(\theta) \hat{A}_t, \\
    & \quad \operatorname{clip}\big(\rho_t(\theta), 1-\epsilon, 1+\epsilon\big) \hat{A}_t \bigg) \\
    & - \beta \mathbb{D}_{\text{KL}}(\pi_\theta \| \pi_{\text{ref}}) \Bigg] ,
\end{split}
\end{equation}
where $\rho_t(\theta)$ is the importance sampling ratio $
\rho_t(\theta)=\frac{\pi_\theta\left(a_t \mid s_t\right)}{\pi_{ref}\left(a_t \mid s_t\right)}$, $\pi_{ref}$ is a fixed reference policy and $\hat{A}_t$ is the estimated advantage function at time step $t$, which will be detailed later.

Here, we employ both the PPO-clip mechanism and a KL divergence penalty to constrain the step size of each model update. We note that this combination of mechanisms have been extensively used and proven successful in RLHF (PPO-like algorithms) for language models~\cite{ouyang2022training,rafailov2023direct,shao2024deepseekmath}, leading us to believe this setup is well-justified.

\subsubsection{Reference-anchored Deterministic Baseline}
\label{subsubsec:learning-free-deterministic-baseline}

In order to compute the advantage function $\hat{A}$ required by the PPO objective, we first introduce Generalized Advantage Estimation (GAE)~\cite{schulman2015high} as the framework for advantage calculation. 
A critical challenge in applying RL to RAG is the high variance and sparsity of LLM-based rewards, which makes training a separate value network (critic) notoriously difficult.
To address this, we introduce a reference-anchored deterministic baseline for defining $V(s_t)$.

\paragraph{GAE in Objective Function}
We use GAE to compute the raw advantage function $A_t$, which is then normalized to obtain $\hat{A}_t$.
GAE is a widely adopted method for effectively estimating the advantage function by balancing bias and variance, which is also used in PPO.
The TD residual from time step $t$ to $t+1$ is defined as
\begin{equation}
\label{eq:td-computation}
\delta_t=r_t+\gamma V(s_{t+1})-V(s_t).
\end{equation}
$A_t$ is then calculated as (for an episode of length $k$):

\begin{equation}
\label{eq:advantage-computation}
\begin{aligned}
A_{t} &= \delta_{t} + (\gamma\lambda)\delta_{t+1} + \dots + (\gamma\lambda)^{k-t-1}\delta_{k-1} \\
&= \sum_{j=t}^{k-1} (\gamma\lambda)^{j-t}\delta_{j} ,
\end{aligned}
\end{equation}
where the parameter $\lambda \in [0, 1]$ controls the bias-variance trade-off in the advantage estimation. A value of $\lambda =0$ would result in using only the single-step TD residual $\delta _t$, which has low bias but high variance. Conversely, $\lambda =1$ would correspond to a Monte Carlo estimate over the rest of the episode, which has high bias but lower variance.
By using a value between 0 and 1, GAE provides a fine-grained control to reduce variance while keeping bias at an acceptable level, which is crucial for stable policy updates.
We explain the value of $\lambda$ in the Appendix~\ref{sec:training-details}.

\paragraph{Limitations of Standard PPO Critics}
Our task exhibits certain characteristics that differ from typical RL modeling in RLHF. In our definition, the episode length $k$ is relatively short, and the reward signal computed by $R_{lm}$ can be highly discrete and exhibit large variance. We found in practice that training a separate critic network to fit $V(s_t)$ was ineffective. Critic models based on deep networks tend to predict numerically continuous values, making it difficult for them to accurately capture the value associated with LLM Reader's evaluations, especially for Factual QA tasks. This can lead to a decline in policy performance.

\paragraph{Reference Policy Rollout as a Deterministic Baseline}
Therefore, departing from traditional PPO and its classic applications in RLHF for LLMs, we adopt a deterministic baseline approach that uses actual observed values. This involves generating a reference trajectory using the reference model, which is then evaluated by the LLM Reader and $R_{lm}$ in the environment. The value of being in state $s_t$ (at step $t$ of the agent's trajectory), denoted $V(s_t)$, is defined as the reward obtained by the reference policy $\pi_{ref}$ executing a greedy sequence of $t$ actions starting from the initial state $s_1$.
Specifically, for each step $j=1, \ldots, t$ in this reference rollout, a document is greedily selected according to $\pi_{ref}$, which is described as action $a_{j}^\prime$:
\begin{equation}
\label{eq:pi_ref-rollout}
a_{j}^\prime = \underset{a \in \mathcal{A}(s_j^{\prime})}{\operatorname{argmax}} \, \pi_{ref}(a \mid s_j^{\prime}), (\text {select document} \, d_{c_j}^\prime )
\end{equation}
where $s_1^{\prime}, \ldots, s_t^{\prime}$ is the path generated by these greedy selections by $\pi_{ref}$, with its initial state $s_1^{\prime}=s_1=\left\{d_1, \ldots, d_N\right\}$. Let $d_{c_j}^\prime$ be the document selected by the reference model at its $j$-th step. The response is generated using these selected documents $\{d_{c_1}^\prime, \ldots, d_{c_t}^\prime\}$
\begin{equation}
\text {response}^\prime _{t}=\text {Reader}_{\text{LLM}}(\text {INST}, q, d_{c_1}^\prime, \ldots, d_{c_t}^\prime).
\end{equation}
The baseline value is then computed as
\begin{equation}
\label{eq:V(s_t)-defenition}
V(s_t)=R_{lm}(\text {ans}, \text {response}^\prime _{t}) .
\end{equation}

\paragraph{Advantages of Reference-Anchored Optimization}

By anchoring the time-dependent baseline $V(s_t)$ to the deterministic greedy performance of the reference policy $\pi_{ref}$, we avoid the complexities of training a critic network and the heavy reliance on human preference data typically required in RLHF.
This approach artfully integrates off-policy insights into PPO's on-policy framework, leveraging the typically strong reference reranker as a stable anchor for advantage estimation. By incorporating this reference-based value into the advantage calculation—alongside the importance sampling ratio and KL divergence penalty—we ensure that policy updates are robustly guided towards outperforming $\pi_{ref}$ while mitigating the risk of significant performance degradation during exploration.

\subsubsection{Training Stabilization Strategies}
\label{subsubsec:training-stabilization}

Even with the deterministic baseline, the raw advantage values $A_t$ can exhibit significant variance. To further stabilize the policy gradient updates, we apply advantage normalization:
\begin{equation}
\label{eq:advantage-mean-std-window}
\hat{A}_t = \frac{A_t - \operatorname{mean}(\mathbf{A}_{\text{window}})}{\operatorname{std}(\mathbf{A}_{\text{window}}) + \epsilon_{\text{norm}}} ,
\end{equation}
where statistics are computed over a sliding window of recent episodes, and $\epsilon_{\text{norm}} = 10^{-8}$.

Additionally, for the KL penalty term in the objective function, we employ the estimator from~\cite{ouyang2022training}:
\begin{equation}
\mathbb{D}_{\text{KL}}(\pi_\theta \| \pi_{\text{ref}}) = \frac{\pi_{\text{ref}}(a_t \mid s_t)}{\pi_\theta(a_t \mid s_t)} - \log \frac{\pi_{\text{ref}}(a_t \mid s_t)}{\pi_\theta(a_t \mid s_t)} - 1 .
\end{equation}

We provide the pesudo codes of \methodname in Algorithm~\ref{alg:pseudo_codes} of Appendix~\ref{sec:pesudo_codes}.

\section{Experiments}
\label{sec:experiments}

\subsection{Experimental Settings}
\label{subsec:experimental-settings}

\paragraph{Datasets}
AmbigNQ~\cite{min-etal-2020-ambigqa} is a disambiguated version of Natural Questions (NQ)~\cite{kwiatkowski2019natural}. AmbigNQ aims to elicit unique, definitive answers by providing more precise question formulations or contexts. 
We use its full training set and evaluate on validation set.
HotpotQA~\cite{yang-etal-2018-hotpotqa} is a widely used multi-hop question answering (multi-hop QA) benchmark dataset, where questions typically require reasoning over information combined from multiple documents. We selected HotpotQA because multi-hop reasoning scenarios place higher demands on the collaborative capabilities of the various modules within RAG systems. 
We use full training set and evaluate on validation set.

\paragraph{Baselines}
To evaluate \methodname, we establish several key comparisons. Our primary baseline is the base reranker model gte-multilingual-reranker-base~\cite{zhang2024mgte} (reported as ``gte reranker'') without \methodname's reinforcement learning optimization, providing a direct ablation of our training paradigm's impact.
We further benchmark against a diverse set of off-the-shelf alternatives. Specifically, we include \textbf{encoder-only} models like jina-reranker-v2-multilingual~\cite{JinaRerankerv2} and bge-reranker-base~\cite{bge_embedding}, alongside the \textbf{decoder-only} model Qwen3-reranker-0.6B~\cite{qwen3embedding} (reported as ``qwen3 reranker'').
For broader context on RAG system performance, we also reference advanced RAG methods like FLARE~\cite{jiang2023active} and DRAGIN~\cite{su-etal-2024-dragin} (with results cited from DeepRAG~\cite{guan2025deeprag} for fair comparison).
These methods typically employ strategies like confidence-based adaptive retrieval, representing orthogonal approaches to RAG optimization rather than direct reranker baselines.

\paragraph{Implementation Details}
We conduct all experiments using the preprocessed knowledge corpus provided by DPR~\cite{karpukhin-etal-2020-dense}, where articles are segmented into 100-word passages.
Our retrieval pipeline employs BM25~\cite{INR-019} to initially fetch $100$ candidate documents, which are subsequently processed by the evaluated reranking models (e.g., gte, jina, and qwen3).
From these reranked candidates, we select the top-$k$ passages ($k=3$ for HotpotQA and $k=5$ for AmbigNQ) to serve as input for the Qwen2.5-7B~\cite{team2024qwen2} LLM reader.
Regarding the optimization objective, we follow previous work~\cite{ma-etal-2023-query} by adopting a unified reward function $R_{lm} = \text{EM} + \lambda _f \text{F1} + \lambda _h \text{Hit}$ for both datasets, with hyperparameters set to $\lambda _f = 1.0$ and $\lambda _h = 1.0$.
The Hit metric is defined to be $1$ if the ground-truth answer appears in the generated response and $-1$ otherwise.
Further training details including fair comparison discussions are provided in Appendix~\ref{sec:training-details}.

\begin{table}[t]
    \centering
    \small  
    \setlength{\tabcolsep}{3.5pt} 
    \begin{tabular}{lcccc}
        \toprule
        \multirow{2}{*}[-0.5ex]{Methods} & \multicolumn{2}{c}{HotpotQA} & \multicolumn{2}{c}{AmbigNQ} \\
        \cmidrule(lr){2-3} \cmidrule(lr){4-5}
         & EM & F1 & EM & F1 \\
        \midrule
         LLM Reader only & 17.80 & 26.85 & 3.55 & 17.80 \\
        LLM + CoT & 21.34 & 31.17 & 10.69 & 27.85 \\
        \midrule
         FLARE & 23.40\textsuperscript{*} & 32.06\textsuperscript{*} & 25.62 & 37.78 \\
         DRAGIN & 16.70\textsuperscript{*} & 24.60\textsuperscript{*} & 18.33 & 30.09 \\
        \midrule
         Naive RAG (BM25) & 24.23 & 35.42 & 20.33 & 32.18 \\
        \midrule
  
         \hspace{2mm}+ qwen3 reranker & 17.03 & 25.71 & 31.17 & 42.87 \\
         \hspace{2mm}+ bge reranker & 28.06 & 40.46 & 31.37 & 45.01 \\
         \hspace{2mm}+ jina reranker & 28.09 & 40.65 & 32.27 & 45.53 \\
         \hspace{2mm}+ gte reranker & 28.24 & 41.23 & 35.51 & 48.55 \\
        \midrule
         \hspace{2mm}+ qwen3 reranker (\methodname) & 18.43 & 27.73 & 32.52 & 45.17 \\
         \hspace{2mm}+ bge reranker (\methodname) & 28.12 & 40.76 & 32.37 & 45.87 \\
         \hspace{2mm}+ jina reranker (\methodname) & 29.26 & 42.16 & 32.86 & 46.20 \\ 
         \hspace{2mm}+ gte reranker (\methodname) & \textbf{30.03} & \textbf{43.22} & \textbf{36.56} & \textbf{49.67} \\        
        \bottomrule
    \end{tabular}
    \caption{Main experimental results on the HotpotQA and AmbigNQ datasets. \textsuperscript{*} adapts from previous work DeepRAG~\protect\cite{guan2025deeprag}.}
    \label{tab:main_results}
\end{table}

\subsection{Main Results}
\label{subsec:main-results}

The results in Table~\ref{tab:main_results} demonstrate that while standard pre-trained rerankers significantly improve upon Naive RAG, applying the \methodname paradigm yields decisive further gains.
Specifically, the \methodname-optimized gte reranker boosts F1 scores by 1.99 on HotpotQA and 1.12 on AmbigNQ, with similar consistency observed across other architectures like jina.
Crucially, our results demonstrate that \methodname is effective across diverse model architectures, yielding consistent gains for both \textbf{encoder-only} and \textbf{decoder-only} rerankers.
Statistical significance tests, detailed in Appendix~\ref{sec:statistical-significance-tests}, confirm that these improvements are statistically significant.
These results not only verify that our RL-based optimization successfully aligns retrieval with downstream generation needs but also establish \methodname as highly competitive against advanced RAG strategies like FLARE and DRAGIN.

\subsection{Ablation Study}
\label{subsec:ablation-study}

\begin{figure}[t]
    \centering
    \includegraphics[width=\linewidth]{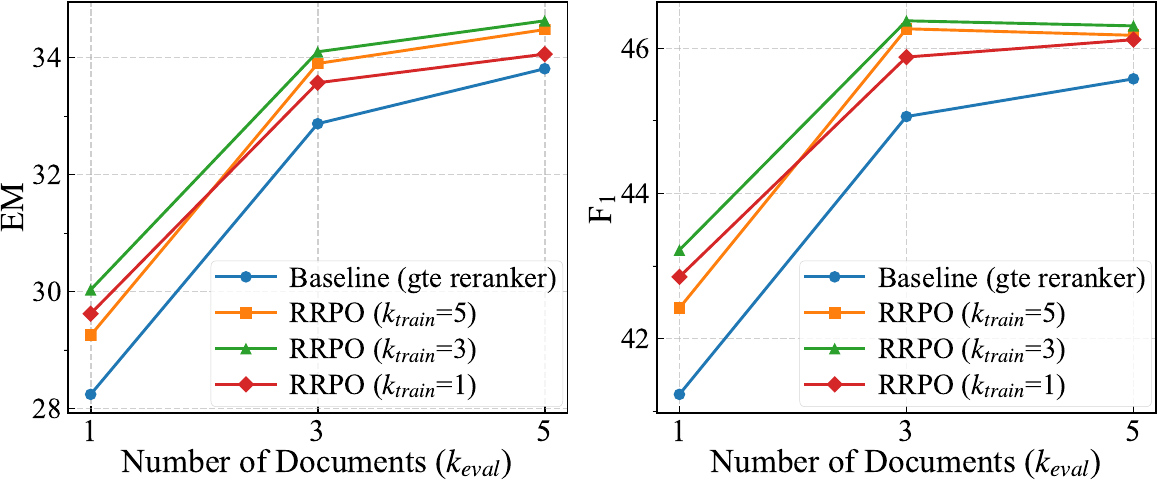}
    \caption{Ablation Experiments on HotpotQA.}
    \label{fig:ablation_topk_hotpotqa}
\end{figure}

Figure~\ref{fig:ablation_topk_hotpotqa} investigates the impact of training interaction depth ($k_{train} \in \{1, 3, 5\}$) on HotpotQA.
All \methodname variants consistently outperform the non-RL baseline, demonstrating robustness.
Notably, $k_{train}=3$ yields optimal results, balancing sufficient context against the noise and attention diffusion of larger sets ($k_{train}=5$) and the limited scope of single-document training ($k_{train}=1$).

\subsection{Extensive Analysis on Generalizability and Robustness}
\label{sec:extensive_analysis}

In this section, we provide a comprehensive analysis demonstrating that \methodname learns a robust, transferable, and highly efficient document utility policy.
Going beyond standard benchmarks, our experiments validate five key properties of the proposed framework:
(1) it generalizes effectively to structurally complex multi-hop reasoning scenarios;
(2) it transfers seamlessly across diverse LLM readers (including closed-source models) without fine-tuning;
(3) it outperforms state-of-the-art list-wise rerankers by explicitly optimizing for reader utility;
(4) it yields orthogonal gains when integrated into advanced RAG pipelines;
and (5) it maintains high label efficiency even when distilled from smaller supervisors.

\begin{figure*}[t]
    \centering
    \includegraphics[width=0.95\linewidth]{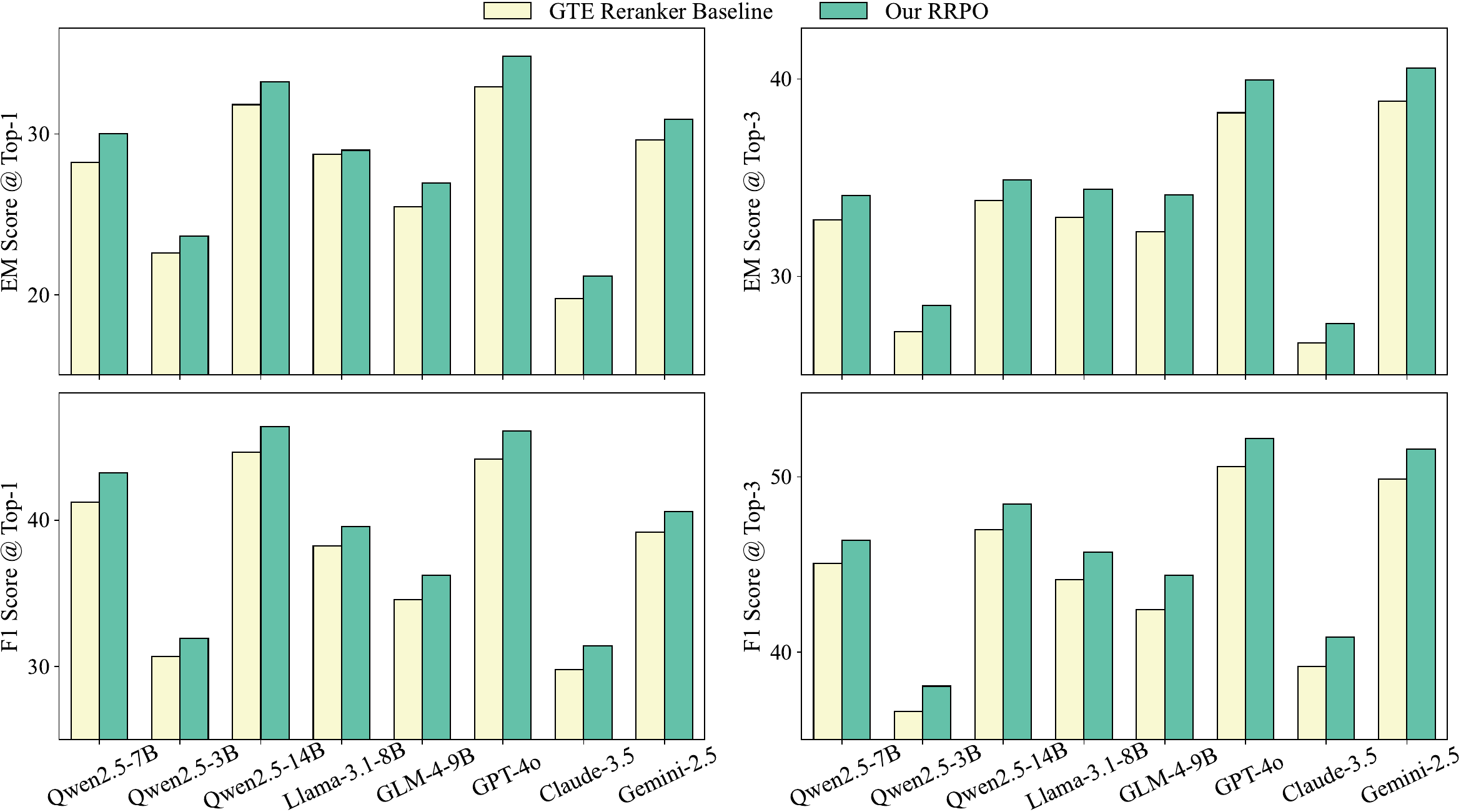}
    \caption{Generalization of \methodname-trained reranker to Various LLM Readers on HotpotQA.}
    \label{fig:generalization_llm_reader}
\end{figure*}

\textbf{\methodname demonstrates robust generalization to structurally complex multi-hop datasets.}
Instead of limiting our evaluation to simple pattern matching, we extend the scope to 2WikiMultiHopQA~\cite{ho-etal-2020-constructing} and MusiQue~\cite{DBLP:journals/tacl/TrivediBKS22}, datasets that necessitate rigorous information integration across disconnected documents.
Using the identical training configuration as the HotpotQA experiments, Table~\ref{tab:generalization_2wiki} reports the performance in the top-5 context setting (with comprehensive results in Appendix~\ref{sec:full_2wiki}).
\methodname consistently outperforms the strong gte reranker baseline across these benchmarks.
Crucially, this performance gain on complex multi-hop datasets confirms that our method does not merely overfit to specific query patterns.
Instead, it successfully learns to identify the cohesive chain of evidence required for multi-hop reasoning, validating its semantic robustness beyond the training domain.

\begin{table}[t]
    \centering
    \small
    \begin{tabular}{lll}
        \toprule
        Methods & {EM} & {F1} \\
        \midrule
        Naive RAG (BM25) & 26.11 & 30.40 \\
        \quad + gte reranker & 26.87 & 31.42 \\
        \quad + gte reranker (\methodname) & \textbf{27.68} & \textbf{32.48} \\
        \bottomrule
    \end{tabular}
    \caption{Performance of \methodname on 2WikiMultiHopQA.}
    \label{tab:generalization_2wiki}
\end{table}

\textbf{The learned ranking policy proves highly transferable across diverse LLM architectures and scales.}
We investigate whether \methodname captures a universal criterion for document utility regardless of the downstream reader.
As illustrated in Figure~\ref{fig:generalization_llm_reader}, our method demonstrates strong ``plug-and-play'' potential across a wide spectrum of models without any reader-specific fine-tuning.
First, the gains consistently transfer across model scales, benefiting both smaller (3B) and larger (14B) variants of the Qwen family. 
More importantly, the policy proves effective across differing architectures, extending benefits to Llama-3.1 and GLM-4.
This cross-architecture transferability suggests that \methodname relies on intrinsic semantic information density rather than model-specific parametric knowledge.
Finally, the method acts as a genuinely reader-agnostic filter, enhancing even proprietary, closed-source models such as Gemini-2.5-Flash, Claude-3.5-Sonnet, and GPT-4o.

\begin{table}[t]
    \centering
    \small
    \setlength{\tabcolsep}{3.5pt} 
    \begin{tabular*}{\linewidth}{@{\extracolsep{\fill}}lcccc}
        \toprule
        \multirow{2}{*}[-0.5ex]{Methods} & \multicolumn{2}{c}{HotpotQA} & \multicolumn{2}{c}{AmbigNQ} \\
        \cmidrule(lr){2-3} \cmidrule(lr){4-5}
         & EM & F1 & EM & F1 \\
        \midrule
        BM25 + gte reranker & 32.87 & 45.06 & 40.26 & 51.36 \\
        BM25 + RankZephyr & 32.40 & 44.49 & 40.06 & 51.58 \\
        \midrule
        BM25 + gte reranker (\methodname) & \textbf{34.10} & \textbf{46.38} & \textbf{41.11} & \textbf{52.18} \\
        \bottomrule
    \end{tabular*}
    \caption{Comparison of \methodname and the list-wise LLM reranker RankZephyr.}
    \label{tab:comparison-with-rankzephyr}
\end{table}

\textbf{\methodname outperforms state-of-the-art list-wise rerankers by explicitly optimizing for reader utility.}
To position our method within the current landscape, we compare it against RankZephyr~\cite{pradeep2023rankzephyr}, a representative 7B list-wise LLM reranker known for its GPT-4-level performance.
As shown in Table~\ref{tab:comparison-with-rankzephyr}, \methodname achieves superior performance compared to RankZephyr in the top-3 context setting across both HotpotQA and AmbigNQ.
While list-wise approaches leverage the general reasoning capabilities of LLMs to sort documents, our results indicate that they may not be optimally aligned with the specific needs of a reader in RAG.
In contrast, \methodname's policy—distilled through \methodname—is explicitly optimized to maximize reader performance, allowing it to filter for utility more effectively than general-purpose ranking prompts.

\textbf{\methodname provides orthogonal improvements to query expansion techniques within advanced RAG frameworks.}
Retrieval pipelines often employ query expansion (e.g., Query2Doc~\cite{wang-etal-2023-query2doc}) to address low recall. We explore whether our method contributes value beyond merely compensating for weak initial retrieval.
Table~\ref{tab:fusion-results} reveals a strictly tiered improvement structure.
Even atop the strong baseline established by Query2Doc, \methodname delivers significant additional gains.
This confirms that the two components play orthogonal roles: while Query2Doc expands the search boundary to enhance Recall, \methodname optimizes the context utility to enhance Precision.
Consequently, our method proves essential even when initial retrieval quality is already high, serving as a critical refinement stage in sophisticated RAG pipelines.

\begin{table}[t]
    \centering
    \small 
    \setlength{\tabcolsep}{3.5pt} 
    \begin{tabular}{lcccc}
        \toprule
        \multirow{2}{*}[-0.5ex]{Methods} & \multicolumn{2}{c}{HotpotQA} & \multicolumn{2}{c}{AmbigNQ} \\
        \cmidrule(lr){2-3} \cmidrule(lr){4-5}
         & EM & F1 & EM & F1 \\
        \midrule
        Naive RAG (BM25) & 24.23 & 35.42 & 20.33 & 32.18 \\
        \hspace{1mm}+ Query2Doc & 28.79 & 41.78 & 36.06 & 50.85 \\
        \hspace{3mm}+ gte reranker & 28.87 & 42.06 & 38.76 & 53.14 \\
        \hspace{3mm}+ gte reranker (\methodname) & \textbf{30.51} & \textbf{44.10} & \textbf{39.81} & \textbf{54.10} \\
        \bottomrule
    \end{tabular}
    \caption{Performance of \methodname within the Query2Doc-enhanced pipeline on HotpotQA and AmbigNQ.}
    \label{tab:fusion-results}
\end{table}

\textbf{The framework maintains high label efficiency even when distilled from smaller, lightweight supervisors.}
Finally, we examine the cost-efficiency of our approach by training a reranker using a smaller, potentially ``noisier'' supervisor (Qwen2.5-3B), denoted as \methodname-3B.
Table~\ref{tab:partial-results-on-smaller-llm-readers} reports its performance (with full details in Appendix~\ref{sec:experiments-on-smaller-llm-readers}).
Remarkably, \methodname-3B delivers consistent improvements.
This resilience to supervisor noise suggests that the core signals for document utility are prominent enough to be captured even by lightweight models.
This finding highlights the label efficiency of our approach, demonstrating that a robust reranking policy can be distilled without relying on computationally expensive, large-scale supervisors.

\begin{table}[t]
    \centering
    \small
    \begin{tabular*}{\linewidth}{@{\extracolsep{\fill}}llcc}
        \toprule
        LLM Reader & Methods & EM & F1 \\
        \midrule
        \multirow{2}{*}{Qwen2.5-3B} 
        & + GTE (Baseline) & 22.59 & 30.69 \\
        & + GTE (\methodname-3B) & \textbf{23.36} & \textbf{31.56} \\
        \bottomrule
    \end{tabular*}
    \caption{Performance of \methodname trained with a smaller LLM Reader (Qwen2.5-3B).}
    \label{tab:partial-results-on-smaller-llm-readers}
\end{table}

\section{Conclusion}
\label{sec:conclusion}

In this work, we propose \methodname to address the fundamental misalignment between static retrieval metrics and the dynamic needs of LLM readers in RAG systems.
By formulating reranking as a sequential decision-making process anchored by a deterministic baseline, we successfully stabilize the reinforcement learning training, eliminating the dependency on expensive human annotations or unstable critic networks.
Our extensive experimental results offer a critical insight:
optimizing for context utility yields significantly better RAG performance than merely optimizing for semantic relevance.
Beyond performance gains, \methodname demonstrates remarkable robustness---it generalizes across diverse LLM readers (including proprietary models like GPT-4o) and distills effective policies even from smaller, noisy supervisors.
This establishes \methodname not just as a performance booster, but as a scalable, cost-efficient paradigm for constructing task-aware RAG systems.

\section*{Limitations}

As a reranking model, \methodname operates on a candidate set of documents provided by an initial retrieval module. If the initial retriever fails to retrieve a sufficient number of relevant documents (i.e., low recall in the top-$N$ candidates), the ability of \methodname to improve the final RAG performance will be inherently limited, regardless of how well it reorders the provided candidates.

\section*{Ethics Statement}

This work does not pose any ethical issues. All the data and models used in this paper are publicly available and are used under following licenses: Creative Commons BY 4.0 License, MIT License, Apache license 2.0.

\section*{Acknowledgements}

This work was supported by the National Natural Science Foundation of China under Grant No. 62476134 and No. 62376120.

\bibliography{refs}

\clearpage

\appendix
\section{Pseudo Codes of \methodname}
\label{sec:pesudo_codes}
Here we present the pseudo codes for our \methodname algorithm.

\vspace{10pt}
\hrule height 0.8pt 
\vspace{2pt}

\refstepcounter{algorithm}
\noindent\textbf{Algorithm \thealgorithm:} \methodname: RL for Reranker Optimization
\label{alg:pseudo_codes}
\vspace{2pt}

\hrule height 0.5pt 
\vspace{2pt}
\begin{algorithmic}[1]
\State \textbf{Input:} Initial retriever, Reader$_{\text{LLM}}$, ground truth answers `ans', instruction `INST', eval metric $R_{lm}$.
\State \textbf{Hyperparameters:} $k$ (docs to select), $\alpha$ (learning rate), PPO params ($\epsilon, \beta$), GAE params ($\gamma, \lambda$), $M_{iter}$ (iterations), $E_{ppo}$ (PPO epochs), $B_{size}$ (batch size).
\State Initialize policy network (reranker) $f_\theta$; Fixed reference policy $\pi_{ref}$ (from $f_{ref}$).
\State Initialize experience buffer $\mathcal{D}$.

\For{iteration $m = 1$ to $M_{iter}$}
  \State Clear buffer $\mathcal{D}$.
  \For{each training query $q$ in a batch of $B_{size}$}
    \State $D_{cand} \leftarrow \text{Retriever}(q)$; $s_1 \leftarrow D_{cand}$; $L_{selected} \leftarrow []$.
    \State $\tau_{query\_data} \leftarrow []$. \Comment{Temporary storage for the current query's trajectory}

    \Statex \hspace*{\algorithmicindent} {\textit{1. Agent Rollout \& Reward Calculation}}
    \For{$t = 1$ to $k$}
     \State Select document $d_{c_t}$ from $s_t$ using policy $\pi_\theta(\cdot|s_t; q)$ (derived from $f_\theta$, Eq.~\eqref{eq:f_theta}).
     \State $L_{selected} \leftarrow L_{selected} \cup \{d_{c_t}\}$; $s_{t+1} \leftarrow s_t \setminus \{d_{c_t}\}$.
     \State $response_t \leftarrow \text{Reader}_{\text{LLM}}(\text{INST}, q, L_{selected})$.
     \State $r_t \leftarrow R_{lm}(\text{ans}_q, \text{response}_t)$ (Eq.~\eqref{eq:compute_reward}).
     \State Add $(s_t, d_{c_t}, r_t, \log \pi_\theta(d_{c_t}|s_t))$ to $\tau_{query\_data}$.
    \EndFor

    \Statex \hspace*{\algorithmicindent} {\textit{2. Learning-free Deterministic Baseline Calculation}}
    \For{$t = 1$ to $k$} \Comment{For each step $t$ in the agent's trajectory}
     \State $L'_{sel\_for\_V_t} \leftarrow$ docs selected by $t$-step greedy rollout of $\pi_{ref}$ on $D_{cand}$ for $q$ (Eq.~\eqref{eq:pi_ref-rollout}).
     \State $V_t \leftarrow R_{lm}(\text{ans}_q, \text{Reader}_{\text{LLM}}(\text{INST}, q, L'_{sel\_for\_V_t}))$ (This is $V(s_t)$ in Eq.~\eqref{eq:V(s_t)-defenition}).
     \State Augment $t$-th step data in $\tau_{query\_data}$ with its corresponding $V_t$.
    \EndFor

    \Statex \hspace*{\algorithmicindent} {\textit{3. Advantage Estimation (GAE)}}
    \State Compute GAE advantages $A_1, \dots, A_k$ for $\tau_{query\_data}$ using $r_t, V_t, \gamma, \lambda$ (Eqs.~\eqref{eq:td-computation}-\eqref{eq:advantage-computation}
    \State Store trajectory (now including $A_t$) from $\tau_{query\_data}$ into $\mathcal{D}$.
  \EndFor \Comment{End of batch collection}

  \Statex \hspace*{\algorithmicindent} {\textit{4. Advantage Normalization}}
  \State Normalize all advantages $A_t$ in $\mathcal{D}$ to get $\hat{A}_t$ (Eq.~\eqref{eq:advantage-mean-std-window}, e.g., using mean/std over $\mathcal{D}$).

  \Statex \hspace*{\algorithmicindent} {\textit{5. Policy Optimization (PPO)}}
  \State Update policy parameters $\theta$ for $E_{ppo}$ epochs by optimizing PPO objective (Eq.~\eqref{eq:ppo-objective}) using mini-batches from $\mathcal{D}$ (which contains $s_t, d_{c_t}, \hat{A}_t, \log \pi_\theta(d_{c_t}|s_t)_{old}$, and uses $\pi_{ref}$).
\EndFor
\State \textbf{Output:} Optimized reranker $f_\theta$.
\end{algorithmic}
\vspace{2pt}
\hrule height 0.8pt
\vspace{10pt}

\section{More Implementation Details}

\label{sec:training-details}

\subsection{Training Details}

On the HotpotQA dataset, we trained for 2 epochs with an initial learning rate of $2e-6$. For RL training interactions on this dataset, we used $k_{train}=3$ (i.e., the LLM Reader received 3 documents per interaction). 
On the AmbigNQ dataset, we trained for 4 epochs with an initial learning rate of $1.6e-6$. For RL training interactions on this dataset, we used $k_{train}=5$.
We adjusted some PPO hyperparameters based on the characteristics of the \methodname task. Specifically, in GAE, the discount factor $\gamma=0.99$ and $\lambda=0.95$. For the PPO clipping objective, $\epsilon=0.2$. In our objective function, the KL divergence penalty coefficient $\beta$ was set to 0.1. 
We use AdamW as our Optimizer.
We use 4 * RTX 3090 for training reranker, and 2 * A6000 for deploying LLM Reader using vllm due to accelerate LLM's inference performance. We use torch AMP and Flash Attention for training. We set batch size to 32 using gradient accumulation.

\subsection{Discussion on Experimental Fairness and Baseline Selection}
\label{sec:discussion-on-experimental-fairness-and-baseline-selection}

To ensure the rigor of our experimental conclusions, we strictly control the variables and carefully select baselines based on data availability.

Controlled Evaluation Protocol: As emphasized in the main text, strict variable control is maintained across all experiments. For every query, both the baseline (e.g., gte-reranker) and our optimized model (\methodname) rank the exact same pool of candidate documents (top-$N$) retrieved by BM25. All downstream parameters, including the number of selected documents ($k$) and the LLM Reader configuration, are kept identical. This ensures that any performance gain is exclusively attributable to the reordering policy learned by \methodname.

Why SFT is Not Included as a Baseline: A direct comparison with Supervised Fine-Tuning (SFT) is methodologically challenging in our specific open-domain setting for two reasons:

Absence of Gold Labels: SFT relies on high-quality, passage-level relevance annotations (e.g., query-passage pairs labeled as positive/negative). However, the datasets used in our experiments (HotpotQA and AmbigNQ within the DPR retrieval setup) provide ground-truth answers but do not inherently provide gold-standard ranking labels for the retrieved top-100 candidates.

Domain Mismatch Risks: While one could fine-tune a reranker on external labeled corpora like MS MARCO, evaluating such a model on our target datasets would introduce severe domain distribution shifts, rendering the comparison unfair.

Therefore, comparing against the pre-trained base reranker provides the most direct assessment of \methodname's ability to align retrieval with generation using only the available reader feedback, effectively bypassing the data bottleneck of SFT.

\section{Prompt Template}
\label{sec:prompt_template}

We utilize this prompt template for training and evaluation on both HotpotQA and AmbigNQ dataset.

\begin{tcolorbox}[
    breakable, 
    colback=black!5!white, 
    colframe=black!75!black, 
    arc=2mm, 
    boxrule=0.5pt 
]

\noindent
\textbf{System:}\\
You are a highly capable assistant who can read the provided passages and accurately answer the question below. You only need to answer the information required for the question within a few words, without any additional content. For questions requiring a `yes' or `no' response, answer solely with `Yes' or `No', without providing any additional explanation.

\medskip
\noindent\textbf{User:}\\
passage 1: Steven Peter's father is Steven Pawl.

\medskip
\noindent\textbf{User:}\\
passage 2: Steven Peter's father graduated from Harvard University.

\medskip
\noindent\textbf{User:}\\
Based on these passages, answer the question. question: Where did Steven Pawl graduate from?

\medskip
\noindent\textbf{Assistant:}\\
Harvard University

\medskip
\noindent\textbf{User:}\\
passage 1: \{\{passage\_1\}\}

\medskip
\noindent\textbf{User:}\\
passage 2: \{\{passage\_2\}\}

\medskip
\noindent\textbf{User:}\\
passage 3: \{\{passage\_3\}\} \\

...(the number of passages is determined by $k$.)

\medskip
\noindent\textbf{User:}\\
Based on these passages, answer the question. question: \{\{query\}\}

\end{tcolorbox}

\section{Generalization with Different Prompt Templates}
\label{sec:generalization-with-different-prompt-templates}

Previous sections demonstrated the generalization of our method across different LLM Readers. In this subsection, we investigate a critical question: does the reranker, trained via reinforcement learning with a specific prompt template, become overfitted to that template? To assess robustness, we evaluate performance using two additional templates (Template A and Template B) that differ structurally and stylistically from the default template used in the main experiments.

Template A redefines the system role as an ``information extraction AI'' and groups context under ``Text Snippet'' headers within a single user turn. Template B adopts an ``Objective-Constraints'' format in the system message and positions the question before the ``Supporting Documents.'' Both templates modify the phrasing for conciseness while strictly enforcing Yes/No constraints to maintain consistency with the Qwen2.5-7B model used as the LLM Reader.

The specific structures of Prompt Template A and B are shown below.

\begin{tcolorbox}[
    breakable,
    colback=black!5!white,
    colframe=black!75!black,
    arc=2mm,
    boxrule=0.5pt
]
\noindent\textbf{System:}\\
You are an information extraction AI. Your task is to answer the question using the provided text within a few words. Be concise. For yes/no questions, respond with `Yes' or `No' and nothing more.

\medskip
\noindent\textbf{User:}\\
Here is the text you should use:\\

Text Snippet 1: \{\{passage\_1\}\}

---

Text Snippet 2: \{\{passage\_2\}\}

---

...(passages continue based on $k$)

---

Based on the text snippets above, what is the answer to this question: \{\{query\}\}
\end{tcolorbox}

\begin{tcolorbox}[
    breakable,
    colback=black!5!white,
    colframe=black!75!black,
    arc=2mm,
    boxrule=0.5pt
]
\noindent\textbf{System:}\\
Objective: Answer the user's question within a few words. Constraints: Use the information given in the `Supporting Documents'. For `Yes' or `No' questions, provide only that word as the answer. Be brief.

\medskip
\noindent\textbf{User:}\\
Question: \{\{query\}\} \\

Supporting Documents:

Document [1]: \{\{passage\_1\}\}
             
Document [2]: \{\{passage\_2\}\}

...(passages continue based on $k$)
          
Please provide the answer found in the documents.
\end{tcolorbox}

The experimental results are presented in Table~\ref{tab:generalization_prompt_templates}, with default template results reproduced from Table~\ref{tab:main_results} for comparison.

\begin{table}[htbp]
    \centering
    \small
    \begin{tabular*}{\linewidth}{@{\extracolsep{\fill}}clcc}
        \toprule
        \multirow{2}{*}{Template} & \multirow{2}{*}{Methods} & \multicolumn{2}{c}{HotpotQA} \\
        \cmidrule(lr){3-4}
        & & EM & F1 \\
        \midrule
        
        \multirow{3.5}{*}{Default} 
        & BM25 & 24.23 & 35.42 \\
        & \hspace{2mm}+ GTE & 28.24 & 41.23 \\
        & \hspace{2mm}+ GTE (\methodname) & \textbf{30.03} & \textbf{43.22} \\
        \midrule
        
        \multirow{3.5}{*}{A} 
        & BM25 & 23.08 & 30.80 \\
        & \hspace{2mm}+ GTE & 28.39 & 38.15 \\
        & \hspace{2mm}+ GTE (\methodname) & \textbf{29.89} & \textbf{39.98} \\
        \midrule
        
        \multirow{3.5}{*}{B} 
        & BM25 & 24.17 & 32.17 \\
        & \hspace{2mm}+ GTE & 29.40 & 39.37 \\
        & \hspace{2mm}+ GTE (\methodname) & \textbf{31.07} & \textbf{41.32} \\
        
        \bottomrule
    \end{tabular*}
    \caption{Generalization results on different prompt templates on HotpotQA.}
    \label{tab:generalization_prompt_templates}
\end{table}

As shown in Table~\ref{tab:generalization_prompt_templates}, \methodname demonstrates strong robustness to variations in prompt templates, consistently outperforming both Naive RAG (BM25) and the standard gte reranker across all formats. Notably, while the baseline performance fluctuates due to the LLM's sensitivity to prompt phrasing (e.g., Naive RAG EM scores range from 23.08 to 24.23), \methodname delivers a stable and significant performance boost in every scenario.

This consistency confirms that the reranker has learned to identify intrinsically valuable documents rather than overfitting to the specific linguistic patterns of the training prompt. Such generalization capability is critical for real-world RAG systems, validating \methodname as a plug-and-play component compatible with diverse user-defined prompt structures.

\section{Experimental Results on More Challenging Multi-hop QA Datasets}
\label{sec:full_2wiki}

\begin{table}[t]
    \centering
    \small
    \setlength{\tabcolsep}{3.5pt}
    \begin{tabular}{lcccc} 
        \toprule
        \multirow{2}{*}{Methods} & \multicolumn{2}{c}{top-1} & \multicolumn{2}{c}{top-3} \\ 
        
        \cmidrule(lr){2-3} \cmidrule(lr){4-5} 
         & {EM} & {F1} & {EM} & {F1} \\ 
         
        \midrule
        Naive RAG (BM25) & 23.01 & 28.47 & 25.64 & 30.20 \\
        \quad + gte reranker & 22.87 & 28.99 & \textbf{26.46} & 31.36 \\
        \quad + gte reranker (\methodname) & \textbf{23.06} & \textbf{29.50} & 26.36 & \textbf{31.47} \\ 
        \bottomrule
    \end{tabular}
    \caption{Performance of \methodname on 2WikiMultiHopQA.}
    \label{tab:generalization_2wiki_more}
\end{table}

More experimental results on 2WikiMultiHopQA are reported in Table~\ref{tab:generalization_2wiki_more}.
The results further demonstrate the effectiveness of \methodname on this dataset.

For MuSiQue dataset, due to the task's complexity and the lower recall of the BM25 retriever, the overall performance baseline is not high.

\begin{table}[t]
    \centering
    \small
    \setlength{\tabcolsep}{3.5pt}
    \begin{tabular}{lcccc}
        \toprule
        \multicolumn{1}{l}{\multirow{2}{*}[-0.5ex]{Methods}} & \multicolumn{2}{c}{top-1} & \multicolumn{2}{c}{top-3} \\
        \cmidrule(lr){2-3} \cmidrule(lr){4-5}
         & EM & F1 & EM & F1 \\
        \midrule
        BM25 + gte reranker & 6.00 & 16.33 & 10.22 & 19.22 \\
        BM25 + gte reranker (\methodname) & \textbf{6.21} & \textbf{16.35} & \textbf{10.34} & \textbf{19.84} \\
        \bottomrule
    \end{tabular}
    \caption{Performance of \methodname on Harder MuSiQue.}
    \label{tab:training-on-musique-results}
\end{table}

Despite the limited gains, \methodname still yields positive improvements. This demonstrates that even in challenging scenarios with low-quality initial retrieval, \methodname's alignment with the LLM allows it to filter more valuable information from noisy candidates. This shows the advantage of our single-step optimization even within a complex multi-hop context.

\section{Experimental Results on Smaller LLM Readers}
\label{sec:experiments-on-smaller-llm-readers}

In our main experiments, we choose Qwen2.5-7B as the LLM Reader in RAG pipeline. We now explore whether the smaller LLM Readers can affect the results due to its more noisy reward feedback. We choose Qwen2.5-3B here, and evaluate our results on Qwen2.5-3B, 7B and 14B.

\begin{table}[t]
    \centering
    \small
    \begin{tabular*}{\linewidth}{@{\extracolsep{\fill}}llcc}
        \toprule
        \multirow{2}{*}{LLM Reader} & \multirow{2}{*}{Methods} & \multicolumn{2}{c}{HotpotQA} \\
        \cmidrule(lr){3-4}
        & & EM & F1 \\
        \midrule
        
        \multirow{3}{*}{Qwen2.5-3B} 
        & + GTE & 22.59 & 30.69 \\
        & + GTE (\methodname-3B) & 23.36 & 31.56 \\
        & + GTE (\methodname-7B) & \textbf{23.66} & \textbf{31.94} \\
        \midrule
        
        \multirow{3}{*}{Qwen2.5-7B} 
        & + GTE & 28.24 & 31.94 \\
        & + GTE (\methodname-3B) & 29.47 & 42.65 \\
        & + GTE (\methodname-7B) & \textbf{30.03} & \textbf{43.22} \\
        \midrule
        
        \multirow{3}{*}{Qwen2.5-14B} 
        & + GTE & 31.83 & 44.62 \\
        & + GTE (\methodname-3B) & 33.09 & 45.99 \\
        & + GTE (\methodname-7B) & \textbf{33.26} & \textbf{46.37} \\
        
        \bottomrule
    \end{tabular*}
    \caption{Performance of \methodname with smaller LLM Readers.}
    \label{tab:results-on-smaller-llm-readers}
\end{table}

As the results show, the reranker trained under the guidance of Qwen2.5-3B remains effective, showing a clear performance lift over the baseline, although it is slightly less performant than the reranker trained with the 7B model.

\section{Analysis of Alternative RL Architectures}
\label{sec:analysis-of-alternative-rl-architectures}

To validate the necessity of our sequential formulation and reference-anchored baseline, we compare \methodname{} against two standard RL variants on the AmbigNQ dataset: (1) Listwise Bandit, which models the selection of top-$k$ documents as a single action, and (2) Standard PPO, which utilizes a parameterized value network (Critic) instead of our deterministic reference baseline.

\begin{table}[t]
    \centering
    \small
    \setlength{\tabcolsep}{0pt}
    \begin{tabular*}{\linewidth}{@{\extracolsep{\fill}}lcccc}
        \toprule
        \multirow{2}{*}{Methods} & \multicolumn{2}{c}{top-1} & \multicolumn{2}{c}{top-3} \\
        \cmidrule(lr){2-3} \cmidrule(lr){4-5}
         & EM & F1 & EM & F1 \\
        \midrule
        BM25 + GTE & 35.51 & 48.55 & 40.26 & 51.36 \\
        BM25 + GTE (Listwise Bandit) & 36.21 & 49.06 & 39.81 & 50.95 \\
        BM25 + GTE (Standard PPO) & 35.91 & 48.97 & 39.66 & 51.34 \\
        BM25 + GTE (\methodname) & \textbf{36.56} & \textbf{49.67} & \textbf{41.11} & \textbf{52.18} \\
        \bottomrule
    \end{tabular*}
    \caption{Training on alternative RL architectures on AmbigNQ.}
    \label{tab:training-on-alternative-rl-architectures}
\end{table}

As the results in Table \ref{tab:training-on-alternative-rl-architectures} show, both alternative methods achieve improvements over the static baseline (BM25 + GTE) in Top-1 metrics, confirming the validity of incorporating LLM feedback. However, their performance on Top-3 metrics is inconsistent, often stagnating compared to the baseline. This suggests that standard methods struggle to optimize the utility of the entire list (ranking) while successfully promoting single relevant documents (hitting), likely due to the difficulty of credit assignment in single-action settings or the high variance of parametric critics. In contrast, \methodname{} delivers consistent gains across both Top-1 and Top-3 metrics, demonstrating that our sequential step-wise reward mechanism effectively guides the model to construct a high-utility set of documents rather than merely identifying a single hit.

\section{Statistical Significance Tests}
\label{sec:statistical-significance-tests}

\begin{table}[htbp]
    \centering
    \small
    \begin{tabular}{lcccc}
        \toprule
        \multicolumn{1}{c}{\multirow{2}{*}[-0.5ex]{Notes}} & \multicolumn{2}{c}{HotpotQA} & \multicolumn{2}{c}{AmbigNQ} \\
        \cmidrule(lr){2-3} \cmidrule(lr){4-5}
         & EM & F1 & EM & F1 \\
        \midrule
        Trained Results Avg. & 29.89 & 43.09 & 36.61 & 49.59 \\
        Trained Results Std. & 0.09	& 0.12 & 0.15 & 0.15 \\
        \bottomrule
    \end{tabular}
    \caption{Statistical Significance Tests.}
    \label{tab:statistical-significance-tests}
\end{table}

To rigorously validate that the performance improvements achieved by our \methodname framework are not a result of random chance, we conducted statistical significance tests. We employed a paired t-test to compare the per-query performance of our best-performing model, ``+ gte reranker (\methodname)'', against the primary baseline, ``+ gte reranker''. The test was performed on the F1 scores obtained for each query in the validation sets of both HotpotQA and AmbigNQ.

The null hypothesis ($H_0$) for our test is that there is no statistically significant difference between the mean F1 scores of the two models. We aim to reject this hypothesis by demonstrating a sufficiently low p-value.

The results are illustrated in Table~\ref{tab:statistical-significance-tests}. Our analysis yielded that On HotpotQA and AmbigNQ dataset, the paired t-test resulted in a p-value of $p<0.01$. In both cases, the extremely low p-values are well below the standard significance level of $\alpha =0.05$. Therefore, we reject the null hypothesis. This provides strong statistical evidence that the gains observed from applying \methodname are consistent and meaningful, confirming the effectiveness of our proposed training paradigm.

\section{Training Acceleration}
\label{sec:training_acceleration}
To enhance training efficiency and mitigate the computational overhead of the RL-based RAG pipeline, we implement a comprehensive set of optimization strategies:

\begin{itemize}
    \item \textbf{Kernel Acceleration:} We leverage Flash Attention and Automatic Mixed Precision (AMP) to optimize the reranker, and deploy the LLM reader using vLLM~\cite{kwon2023efficient} to maximize inference throughput.
    
    \item \textbf{Retrieval Caching:} To eliminate redundant retrieval operations, we cache the top-N documents obtained from the initial retriever, reusing them across subsequent training epochs.
    
    \item \textbf{LLM Response Replay:} By enforcing deterministic LLM generation (temperature=0), we construct a replay buffer to store and reuse historical responses. This mechanism is critical, as it reduces the required LLM inference calls by nearly 50\% in later training stages.
    
    \item \textbf{Distributed Training Optimization:} We employ gradient accumulation to increase the effective batch size and fine-tune communication schedules for improved parallel efficiency.
\end{itemize}

\section{Extended Comparison with DPO-based Joint Training Paradigms}
\label{sec:appendix_dynamicrag}

As discussed in our related work (Section \ref{subsec:rl-for-rag}), a growing trend in aligning retrieval with generation involves applying preference optimization (e.g., DPO) to fine-tune models. To further contextualize \methodname within these recent advancements, we provide an extended empirical comparison against DynamicRAG \citep{sun2025dynamicragleveragingoutputslarge}, a representative concurrent work in this direction.

While sharing a similar high-level motivation of utilizing LLM feedback, the two frameworks explore fundamentally different paradigms. DynamicRAG adopts an end-to-end joint training approach, where a 7B LLM serves simultaneously as the reranker and the generator. While this enables deep synergy between components, it inherently introduces substantial computational overhead during both training and inference. In contrast, \methodname focuses on a modular, plug-and-play paradigm. We distill the RL-based alignment strictly into lightweight, encoder-only pointwise models (e.g., gte, jina, bge) with approximately 0.3B parameters, while keeping the downstream LLM generator strictly frozen.

As shown in Table~\ref{tab:dynamicrag_comp}, we present a direct empirical comparison on the HotpotQA dataset. It is worth noting that this constitutes an asymmetric comparison, given DynamicRAG's joint tuning of a 7B parameter space compared to our isolated tuning of a 0.3B reranker. Remarkably, despite this vast difference in scale and the lack of generator fine-tuning, the \methodname-optimized gte model (0.3B) effectively outperforms the 7B DynamicRAG model. Furthermore, when applied to other lightweight encoders, \methodname maintains highly competitive performance. These results demonstrate that our pointwise RL formulation provides a highly efficient alternative, effectively aligning the reranker with reader needs without the steep deployment costs of joint LLM training.

\begin{table}[h]
\centering
\small
\begin{tabular}{lcc}
\toprule
\textbf{Methods} & \textbf{Model Size} & \textbf{EM} \\
\midrule
DynamicRAG (LLaMA-2) & 7B & 29.40\rlap{$^*$} \\
\methodname (bge reranker) & 0.3B & 28.12 \\
\methodname (jina reranker) & 0.3B & 29.26 \\
\methodname (gte reranker) & \textbf{0.3B} & \textbf{30.03} \\
\bottomrule
\end{tabular}
\caption{Performance comparison between lightweight \methodname variants and the LLM-based DynamicRAG on the HotpotQA dataset. $^*$Results directly cited from the original paper.}
\label{tab:dynamicrag_comp}
\end{table}

\end{document}